\pdfoutput=1

\documentclass[11pt]{article}

\usepackage[final]{ACL2023}

\usepackage{times}
\usepackage{latexsym}

\usepackage[T1]{fontenc}

\usepackage[utf8]{inputenc}

\usepackage{microtype}

\usepackage{inconsolata}

\usepackage{microtype}
\usepackage{color}
\usepackage{tabulary,booktabs}
\usepackage{enumitem}
\usepackage{amsmath}
\usepackage{amssymb}
\usepackage{multirow}
\usepackage{adjustbox}
\usepackage{bm}
\usepackage{tikz}
\usepackage{pgf}
\usepackage{pgfplots}
\usetikzlibrary{patterns}
\usepackage{tikz-qtree}
\usepackage{mathrsfs}

\usetikzlibrary{arrows,decorations.pathmorphing,backgrounds,positioning,fit,petri,shapes.misc, arrows.meta,shapes.geometric,decorations.markings,calc,shadows.blur,decorations.pathreplacing,quotes,matrix,shapes.symbols}
\usetikzlibrary{arrows,decorations.pathmorphing,backgrounds,positioning,fit,petri,shapes.misc, arrows.meta,shapes.geometric,decorations.markings,calc,shadows.blur,decorations.pathreplacing,quotes,matrix,shapes.symbols}
\tikzset{
	table/.style={
		first column text width/.code={%
			\tikzset{
				column 1/.style={
					nodes={text width=##1, font=\bfseries}
				},
			}
		},
		first column text width=5em
	}
}

\renewcommand*{\vec}[1]{\mathbf{#1}}

\usepackage{graphicx} 
\graphicspath{{imgs/}} 

\def\Section {\S}

\newcommand{\squishlist}{
\begin{list}{$\bullet$}
{ \setlength{\itemsep}{0pt}
	\setlength{\parsep}{3pt}
	\setlength{\topsep}{3pt}
	\setlength{\partopsep}{0pt}
	\setlength{\leftmargin}{1.5em}
	\setlength{\labelwidth}{1em}
	\setlength{\labelsep}{0.5em} } }

\newcounter{Lcount}
\newcommand{\squishlisttwo}{
\begin{list}{\arabic{Lcount}. }
	{ \usecounter{Lcount}
		\setlength{\itemsep}{0pt}
		\setlength{\parsep}{0pt}
		\setlength{\topsep}{0pt}
		\setlength{\partopsep}{0pt}
		\setlength{\leftmargin}{2em}
		\setlength{\labelwidth}{1.5em}
		\setlength{\labelsep}{0.5em} } }

\newcommand{\squishend}{\end{list} }

\definecolor{fontgray}{RGB}{44, 62, 80}
\definecolor{myred}{RGB}{235, 47, 6} 
\definecolor{summertime}{RGB}{245, 205, 121}
\definecolor{darkgrass}{RGB}{0, 148, 50}
\definecolor{myblue}{RGB}{0, 168, 255}
\definecolor{mygray}{RGB}{158, 158, 158}
\definecolor{puffin}{RGB}{250, 152, 58}
\definecolor{lowpurple}{RGB}{210, 180, 222}
\definecolor{lowblue}{RGB}{102,178,255}
\definecolor{lowred}{RGB}{245, 183, 177}
\definecolor{deeppurple}{RGB}{142, 68, 173}
\definecolor{nephritis}{RGB}{39, 174, 96}

\definecolor{deepblue}{RGB}{41, 128, 185}
\definecolor{shymoment}{RGB}{162, 155, 254}
\definecolor{firstdate}{RGB}{250, 177, 160}
\definecolor{mintleaf}{RGB}{0, 184, 148}
\definecolor{alizarin}{RGB}{231, 76, 60}
\definecolor{soaring}{RGB}{149, 175, 192}
\definecolor{electronblue}{RGB}{9, 132, 227}
\definecolor{pinkgla}{RGB}{0, 184, 148}

\title{Sequence-to-Sequence Pre-training with Unified Modality Masking for Visual Document Understanding}

\author{Shuwei Feng$^\ast$ \qquad Tianyang Zhan\thanks{equal contribution} \qquad Zhanming Jie\\
        \bf{Trung Quoc Luong \qquad Xiaoran Jin}\\
        ByteDance AI Lab\\
        \{shuwei.feng, zhantianyang, allan, trung.luong, xiaoran.jin\}@bytedance.com}

\begin{document}
\maketitle
\begin{abstract}
This paper presents GenDoc, a general sequence-to-sequence document understanding model pre-trained with unified masking across three modalities: text, image, and layout. 
The proposed model utilizes an encoder-decoder architecture, which allows for increased adaptability to a wide range of downstream tasks with diverse output formats, in contrast to the encoder-only models commonly employed in document understanding.
In addition to the traditional text infilling task used in previous encoder-decoder models, our pre-training extends to include tasks of masked image token prediction and masked layout prediction. 
We also design modality-specific instruction and adopt both disentangled attention and the mixture-of-modality-experts strategy to effectively capture the information leveraged by each modality. 
Evaluation of the proposed model through extensive experiments on several downstream tasks in document understanding demonstrates its ability to achieve superior or competitive performance compared to state-of-the-art approaches.
Our analysis further suggests that GenDoc is more robust than the encoder-only models in scenarios where the OCR quality is imperfect.

\end{abstract}

\section{Introduction}
Document understanding is a research topic that involves analyzing, understanding, and reasoning over business documents (e.g., invoices, contracts, financial statements, etc.). 
The topic includes a wide range of tasks such as document image classification~\cite{harley2015evaluation}, layout detection~\cite{zhong2019publaynet}, information extraction~\cite{jaume2019funsd}, table detection~\cite{gao2019icdar}, scene text recognition~\cite{neumann2012real}, etc. 

Pre-training with transformer-based models~\cite{vaswani2017attention} for document image understanding receives significant attention recently~\cite{xu2020layoutlm,appalaraju2021docformer,xu2021layoutlmv2,powalski2021going,hwang2021cost,huang2022layoutlmv3,wang2022lilt} as the pre-trained models are able to achieve remarkable improvement on the above downstream tasks.
Most of existing research efforts focus on pre-training multi-modal transformer encoders~\cite{tan2019lxmert,chen2020uniter,su2020vl,peng2022ernie} to encode the textual, layout, and visual information (e.g., LayoutLM~\cite{xu2020layoutlm,xu2021layoutlmv2,huang2022layoutlmv3}).
They proposed various transformer design (e.g., spatial attention) and pre-training tasks (e.g., image-text contrastive learning) to enable interaction among the modalities.
While these pre-trained encoders achieve great performance on some downstream tasks, encoder-decoder models are more appropriate  for generation tasks such as question answering~\cite{mathew2021docvqa} and more flexible to adapt to different kinds of downstream tasks.
Furthermore, encoder-decoder models are prone to suffer less from imperfect optical character recognition (OCR), which is essential for encoder-only models to achieve good performance on certain tasks (\Section \ref{sec:ablation}).

\citet{powalski2021going} proposed a spatial-aware T5 model~\cite{raffel2020exploring} called Text-Image-Layout Transformer (TILT) to incorporate the layout as spatial bias in the attention mechanism. 
However, their pre-training objective only involves the masked language modeling loss which potentially causes the model to rely solely on the textual modality.

Instead, we add relative disentangled attention~\cite{peng2022ernie} in the encoder for better position understanding and incorporate the mixture-of-modality-experts~\cite{wang2021vlmo} in the decoder to capture the modality-specific information.
We then propose a unified modality masking schema and a unified pre-training loss function for each modality (\Section \ref{sec:pretraining}). 
Specifically, we adopt the standard text infilling task~\cite{lewis2020bart} for textual modality.
Inspired by the masked image modeling in encoder models~\cite{bao2021beit} and image generation~\cite{yu2022scaling,ramesh2022hierarchical}, we use the Vector Quantised-Variational AutoEncoder (VQ-VAE)~\cite{van2017neural} to obtain the image tokens and perform masked image token prediction. 
Finally, we also design a masked coordinate prediction task to enhance the spatial awareness of our encoder-decoder model. 
In order to handle different modalities and provide guidance for downstream tasks, we design specific instructions during both pre-training and fine-tuning (\Section \ref{sec:model_input}).

Our main contribution can be summarized as follows:
\squishlist
\item We propose GenDoc, a general sequence-to-sequence (Seq2Seq) multi-modal transformer for document understanding. 
Such a Seq2Seq model allows us to unify the output format of various pre-training and downstream tasks and exhibits robust performance in the presence of imperfect OCR. 
\item We implement the pre-training of all three modalities (text, visual, and layout) by applying text infilling, masked image token prediction, and masked coordinate prediction, respectively.
We also develop effective designs for modality fusion within the encoder and modality separation within the decoder.
\item We conduct experiments on four document understanding tasks and achieve either state-of-the-art or comparable performance. Furthermore, we perform ablation experiments to demonstrate the significance of each pre-training task and modeling architecture.

\squishend

\section{GenDoc}
GenDoc is a general document understanding model that unifies the training of text, vision and layout modality within a single framework.
Figure \ref{fig:model_architecture} shows the model architecture and the pre-training process for different modalities. 
Our model is a transformer model~\cite{vaswani2017attention} consisting of an encoder with additional layout incorporated and a decoder that features a mixture of  modality experts~\cite{wang2021vlmo} to capture the modality-specific information.
In addition, we also include an image backbone to encode the visual images.

\subsection{Input Representation}
\label{sec:model_input}

\begin{figure*}
\begin{center}
	\includegraphics[width=1.\linewidth]{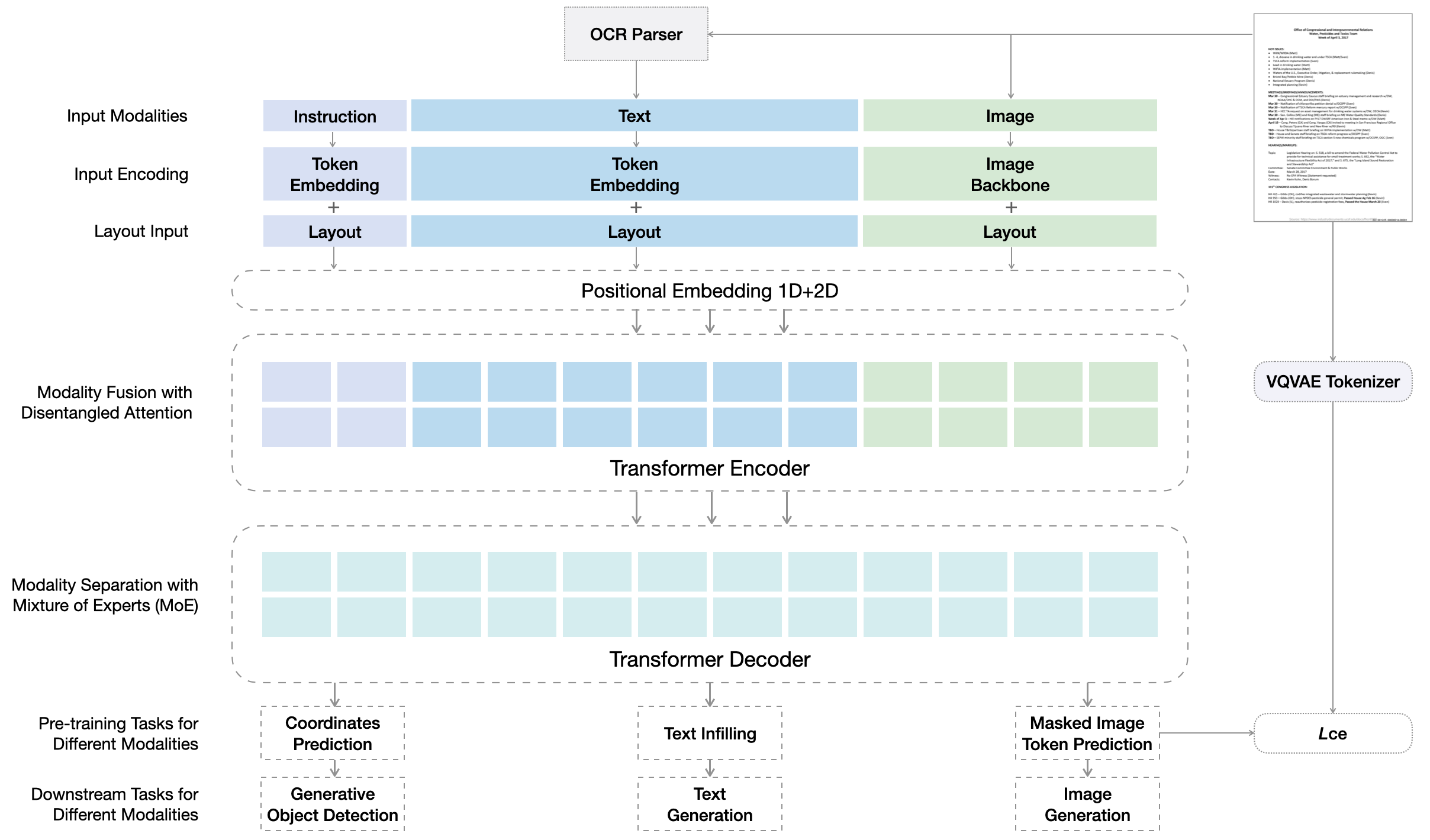}
\end{center}
   \caption{The overview of the proposed model architecture and the pre-training schema.}
\label{fig:model_architecture}
\end{figure*}

As illustrated in Figure \ref{fig:model_architecture}, the input to our model comprises of four primary components: the task-specific instruction, the optional OCR text, the document image and the layout information of text and image patches. 

The instruction and OCR text are tokenized into subwords and then encoded as embeddings. The visual document image is encoded as visual embeddings using an image backbone and the layout information is represented by standalone spatial embeddings. The final input sequence consists of token embeddings and visual embeddings, as well as position embeddings representing 1D and 2D positions integrated.

\paragraph{Instruction}

In order to differentiate between the various pre-training tasks, unique instructions are crafted based on their modalities and appended ahead of other inputs. For example, the instruction ``\textit{What is the complete text for <mask> tokens?}'' is used in the text infilling task, while ``\textit{What are the values for masked image tokens?}'' is used in the masked image token prediction task. During fine-tuning, instructions that are semantically meaningful for each downstream task are designed. These instructions can effectively guide the generation for the corresponding modality and aid the model in adapting to similar tasks for practical application.

\paragraph{OCR Text}

The textual content in a document is obtained through the use of an off-the-shelf OCR parser. Prior research efforts~\cite{huang2022layoutlmv3,powalski2021going} have mainly adopted the use of commercial OCR services such as Microsoft READ API\footnote{https://learn.microsoft.com/en-us/azure/cognitive-services/computer-vision/how-to/call-read-api} or Amazon OCR\footnote{https://aws.amazon.com/textract/}. However, due to budget constraints, we employed our own internal OCR service for pre-training.

For both the instruction and OCR text, we use the subword tokenizer from BART~\cite{lewis2020bart} to tokenize them. As depicted in Figure \ref{fig:model_architecture}, the instruction and OCR text are concatenated together as textual input to the encoder.

\paragraph{Document Image and Visual Tokens}

 The original document image is first resized to a pre-defined size $H \times W \times 3$ where $H$ is the height, $W$ is the width. The image is then passed through the image backbone, a ResNet, to extract features as image patches of size with $h \times w$. These image patches are subsequently flattened as a sequence of vectors for the input of the transformer encoder.

 On the other hand, the document image is quantized into discrete visual tokens during pre-training to build the target for Masked Image Token Prediction task introduced in next section(\Section \ref{sec:pretraining}). A VQ-VAE~\cite{van2017neural} tokenizer is trained on the complete pre-training data, and used to tokenize the images. The tokenized image can be represented as a sequence of tokens $[\bm{z_0},\bm{z_1},...\bm{z_{h\times w}}]$($\bm{z_i} \in { 0, 1, \cdots, \vert \mathcal{V}\vert })$, where $\vert \mathcal{V} \vert =8192$ in our experiments.

\paragraph{Layout}

The layout information is represented by 2D coordinates of each word and image patch. Padding layout is used for the instruction and other special tokens. Since the text extracted from the OCR parser might not appear in a natural language order (typically ordered in a left-to-right and then bottom-to-top manner), a learnable 1D embedding is utilized for positional embeddings. Specifically, separate positional embeddings are utilized for the visual and textual modality (i.e., $\vec{E}_{text\_1d}$ and $\vec{E}_{visual\_1d}$). 
Furthermore, 2D layout representations are also necessary for documents. 
Following the method proposed in \citet{xu2020layoutlm}, 2D layout embeddings are calculated by:
\begin{equation}
	\mathbf{E}_{2d} = \mathbf{E}_{x_1} + \mathbf{E}_{y_1} + \mathbf{E}_{x_2} + \mathbf{E}_{y_2} +\mathbf{E}_{width} + \mathbf{E}_{height}
	\nonumber
\end{equation}
where $(x_1, y_1)$ and $(x_2, y_2)$ represent the top-left and bottom-right coordinates, respectively. 
$width$ and $height$ represent the length of two dimensions. 
The complete tuple $(x_1, y_1, x_2, y_2, width, height)$ uniquely represens a bounding box or a region in a document image.
These layout embeddings are used to represent the 2D coordinate of words and the image patches. 
Formally, our final input embedding representations can be obtained by:
\begin{equation}
	\vec{E} = \left [\vec{E}_t + \vec{E}_{text\_1d}; \vec{E}_v + \vec{E}_{visual\_1d} \right ] + \vec{E}_{2d}
\end{equation}

\subsection{Model Design}
\paragraph{Modality Fusion in Encoder}

In order to reinforce the effectiveness of layout information, we incorporate the disentangled attention into our transformer encoder following previous research efforts~\cite{he2021deberta, peng2022ernie}. 
Specifically, we utilize attentions from content-to-content, content-to-layout in the x-dimension, and content-to-layout in the y-dimension: 
\begin{equation} \label{eq_att}
\begin{split}
A_{ij}^{cc}&=\bm{Q}_{i}^c{\bm{K}_{j}^{c}}^{\intercal} \\
A_{ij}^{cx}&=\bm{Q}_{i}^c{\bm{K}_{\delta_x(i, j)}^{x}}^{\intercal} \\
A_{ij}^{cy}&=\bm{Q}_{i}^c{\bm{K}_{\delta_y(i, j)}^{y}}^{\intercal}  \\
A_{ij} &=A_{ij}^{cc}+A_{ij}^{cx}+A_{ij}^{cy} 
\end{split}
\end{equation}
where $A_{ij}$ is the resulting attention score between position $i$ and $j$; 
$A_{ij}^{cc}$ is the conventional attention score from content query $\boldsymbol{Q}^c$ to content key $\boldsymbol{K}^c$ between position $i$ and $j$; 
$A_{ij}^{cx}$ and $A_{ij}^{cy}$ stand for the disentangled attention scores from content to layout;
$\delta_{\ast}(i,j)$ is the relative distance function between $i$ and $j$, the explicit definition of which can be found in the work of~\cite{he2021deberta}.
Different from previous research efforts, we do not use the attention from content to the 1D position, as the layout order information can be more representative than sequential order for the textual information in the document.

\paragraph{Modality Separation in Decoder}
Some preliminary experiments suggest that our universal decoder could underperform on certain downstream tasks as the pre-training requires the decoder to capture all the information in different modalities (i.e., text, image, and layout). 
Motivated by such an observation, we adopt the mixture-of-modality-experts (MoE)~\cite{wang2021vlmo} strategy in our decoder. 
We use independent feed-forward networks~(FFNs) for different pre-training or downstream tasks.
The activation of specific experts within the decoder is contingent upon the nature of the pre-training or fine-tuning task at hand.
Specifically, the textual expert is used for natural language tasks, visual expert for image token prediction tasks, and layout expert for coordinate prediction tasks.

\subsection{Pre-training Tasks}
\label{sec:pretraining}

In this work, we implement three pre-training tasks utilizing a unified modality masking strategy across three distinct modalities: Text Infilling (TI), Masked Image Token Prediction (ITP), and Masked Coordinate Prediction (CP). Given that an encoder-decoder architecture is employed, all tasks are formulated within a unified generation framework, with the utilization of a shared vocabulary.

\paragraph{Text Infilling}	
Text infilling~\cite{raffel2020exploring,lewis2020bart} is a typical denoising task for pre-training the encoder-decoder architecture. In this work, spans within the text are randomly sampled following the Poisson distribution ($\lambda=3$). The masked text is replaced with \textit{<mask> }tokens, and the corresponding layout is replaced with a padding layout.
 Approximately 30\% of tokens are masked in each document, and we use a special token \textit{<sep>} to join the masked spans as the reconstruction targets.
The objective during pre-training is to recover these \textit{<mask>} spans, similar to the approach employed in T5~\cite{raffel2020exploring}.

\paragraph{Masked Image Token Prediction} 
Similar to text infilling, we build an image denoising method to model image patches inspired by previous studies~\cite{bao2021beit,he2022masked}. Unlike the block-wise masking mechanism used in previous document understanding studies~\cite{huang2022layoutlmv3}, a random masking strategy was adopted, which has been shown to be more robust in conventional visual tasks~\cite{xie2022simmim}. Specifically, 50\% of the image patches produced by the image backbone are replaced with a learnable mask embedding. The pre-training objective is to recover the quantized image token sequence 
$\bm{y}=[\bm{z_0},\bm{z_1},...\bm{z_{h\times w}}]$
of the document image, where each token \bm{$z_i$} is represented by a unique visual token label in the shared vocabulary.

\paragraph{Masked Coordinate Prediction} 

In order to assist the model in comprehending the diverse and complex layouts present within documents, we devise a task named masked coordinate prediction. 
To implement this task, a proportion of spans are selected via Poisson distribution ($\lambda=3$) and their layout coordinates are masked. The span length of the masked segments constitutes 20\% of the total number of tokens in the document.
The objective of this task is to predict the coordinates of the obscured spans, thus providing the model with a means of understanding the layout structure of the document. The resulting sequence comprises a series of coordinates, denoted by $[x_1$, $y_1$, $x_2$, $y_2]$, separated by a special token \textit{<sep>}.

\paragraph{Unified Loss} The training objective for all tasks are unified to minimize the cross-entropy loss: 
\begin{equation} \label{eq_ce}
\mathcal{L}_{ce}=-\sum_{i=1}^{T}\sum_{j=1}^{N_i}\omega_i \log P(\hat{y}|\boldsymbol{x'_i}, y_{1:j-1})
\end{equation}
where $T$ is the number of tasks; $N_i$ is the maximum sequence length for task $i$; $\omega_i$ is the task weight; $\boldsymbol{x'_i}$ is the masked input for task $i$; $\hat{y}$ and $y$ are targets and decoder inputs, respectively.

\section{Experiments}
\label{sec:experiments}
\subsection{Settings and Results}
In order to ascertain the efficacy of our proposed model, a thorough evaluation was conducted on four distinct downstream tasks, namely visual question answering, document layout analysis, form understanding, and document classification.  Experiments were carried out utilizing publicly available benchmarks, including DocVQA~\cite{mathew2021docvqa}, Publaynet~\cite{zhong2019publaynet}, CORD~\cite{Park2019cord}, and RVL-CDIP~\cite{adam2015cdip}. The subsequent sections provide descriptions of the pre-training settings, methods employed for modeling the downstream tasks, and the results obtained. Information regarding the datasets, hyper-parameters and more experimental details can be found in the Appendix \Section \ref{sec:appendix}.

 \begin{table*}
	\centering
	\resizebox{0.85\linewidth}{!}{
		\begin{tabular} {llcccc} \toprule
			& \multirow{2}{*}{\bf Model} & \bf DocVQA & \bf Publaynet  & \bf CORD  & \bf RVL-CDIP\\
			& & \bf  ANLS$\uparrow$& \bf MAP$\uparrow$ & \bf F1$\uparrow$ & \bf Accuracy$\uparrow$ \\ \midrule
			& LayoutLM~\cite{xu2020layoutlm} & - &  - & - & 94.42\\
			& LayoutLMv2~\cite{xu2021layoutlmv2} & 78.08 & - & 94.95 & 95.25 \\
			\textsc{Base} & LILT~\cite{wang2022lilt} & - & - &96.07 &95.68\\
			\textsc{Model} & LayoutLMv3~\cite{huang2022layoutlmv3} & 78.76 & \textbf{95.1} & 96.56 & 95.44 \\ 
			& DocFormer~\cite{appalaraju2021docformer} & - & - & 96.33 & \textbf{96.17} \\
                & DIT~\cite{Li2022dit} & - & 94.5 & - & - \\ 
			& \textbf{GenDoc (Ours)} & \textbf{78.83} & 94.4 & \textbf{96.59} & 93.80 \\ \midrule\midrule
			
			& LayoutLM~\cite{xu2020layoutlm} & -& -& -& 91.90\\
			& LayoutLMv2~\cite{xu2021layoutlmv2} & 83.48& -& 96.01& 95.64 \\
			\textsc{Large}&StructuralLM~\cite{li2021structurallm} & 83.94$\dagger$ & - & - &96.08 \\
			\textsc{Model} & LayoutLMv3~\cite{huang2022layoutlmv3} &83.37  & -& \textbf{97.46}& \textbf{96.56}\\ 
			& DocFormer~\cite{appalaraju2021docformer} & - & - &96.99&95.50\\
			& TILT~\cite{powalski2021going} &87.05$\dagger$ & - &96.33 & 95.52\\
			& \textbf{GenDoc (Ours)} & \textbf{84.49} & - & 96.97 & 94.50 \\ \bottomrule
			
		\end{tabular}
	}
	\caption{Comparison with existing approaches on four document understanding tasks.$\dagger$ indicates they combine the training and validation set for fine-tuning.}
  \label{tab:final_result}
\end{table*}

\paragraph{Pre-training}
To ensure consistency with prior studies, our pre-training is performed using the IIT-CDIP dataset~\cite{david2006iit}, which comprises of 11 million scanned documents.
Textual and layout information are extracted from these documents using an off-the-shelf OCR engine. 

The batch size for each task is set independently, as the understanding of documents relies more heavily on textual information. The base model uses a batch size of $640$ for text infilling, $384$ for masked image token prediction, and $112$ for masked coordinate prediction. The large model employs half the batch size of the base model to save computations. The input images are resized into $448\times448$ ($H \times W$), and the tokenized textual sequences are truncated if they exceed $512$ tokens.
The image backbone is initialized by a standard ResNet-101 model and the transformer encoder-decoder model is initialized by a pre-trained Bart model~\cite{lewis2020bart}. 
The parameters of the visual and layout experts are initialized by the textual expert, while the remaining parameters are randomly initialized.
Both models are pre-trained for $150$k steps with 32 Nvidia-A100 GPUs, and the following experiments are conducted with the pre-trained models.

\paragraph{Document Visual Question Answering}

Document visual question answering is a task that necessitates the model to comprehend document text and images in order to answer specific questions. 

Instead of modeling the DocVQA task as an extractive question answering, as is commonly done in previous work~\cite{xu2021layoutlmv2, huang2022layoutlmv3, li2021structurallm}, we model this task as an abstractive question answering, which benefits from our sequence-to-sequence modeling. 
This approach eliminates the need for non-trivial pre-processing to match the span with the annotated gold answers and enhances the OCR error tolerance of our model.

In the present study, we fine-tune the model on the DocVQA training set, evaluate on the validation set, and obtain the test results by submitting it to the official website\footnote{https://rrc.cvc.uab.es/?ch=17\&com=evaluation\&task=1}. It is observed during the experimentation process that task performance is highly correlated with the quality of OCR. To ensure a fair comparison, we obtain the OCR results through Microsoft Read API without any additional pre-processing. 
Consistent with the pre-training procedure, we include an instruction of ``\textit{What is the answer to the question?}'' preceding the question inputs. The final textual input is formed by concatenating the instruction, question, and document text. The model is fine-tuned for 10 epochs with a batch size of 48. 

We report the ANLS (Average Normalized Levenshtein Similarity) score calculated by the official rated system in Table \ref{tab:final_result}. Without any post-processing, our result outperforms the previous state-of-the-art by $1.1$ on ANLS for the large model and $0.1$ for the base model. These results demonstrate the effectiveness of the GenDoc model in solving document question-answering tasks.

\paragraph{Document Layout Analysis}
Document layout analysis is a task that involves object detection in document images.
PubLayNet is a large dataset that provides abundant annotations of bounding boxes and five document layout categories (i.e. text, list, table, figure, and title). 
We train the model on its training set and report the results on the validation set, as per common practice~\cite{huang2022layoutlmv3}.

Previous approaches use pre-trained models as the feature backbone and add additional modules such as Faster R-CNN~\cite{ren2015fasterrcnn}, Cascade R-CNN~\cite{cai2018cascade}, and FPN~\cite{lin2017fpn} for bounding box regression. In contrast, our model predicts bounding boxes using a language-modeling approach, similar to the coordinate prediction pre-training task. Specifically, the bounding box and its category are formatted as a sequence of discrete tokens, such as [$x_1, y_1, x_2, y_2, class$], and the objective is to maximize the likelihood of the sequence given a document image. To maintain consistency with previous works~\cite{huang2022layoutlmv3,Li2022dit}, we only use images as input for the model.
We use a batch size of $64$ during fine-tuning and train the model for 50 epochs. 

We report the mean average precision (mAP) in Table \ref{tab:final_result}. The result shows our unified model only has a narrow performance gap compared to the strong baseline DiT, which utilizes an extra detection network, Cascade R-CNN. Furthermore, our model down-samples the image to a shorter sequence, as opposed to the patch size of 16 utilized by DiT and LayoutLMv3.
This leads to a reduction in sequence length by $75\%$, resulting in a significant reduction in computation and training costs.

\paragraph{Form Understanding}
Form understanding is a crucial task in the field of document and form information extraction. Commonly used benchmark datasets for this task include FUNSD~\cite{jaume2019funsd} and CORD.
In this work, the Consolidated Receipt Dataset (CORD) is selected as the primary dataset for analysis due to its superior robustness in terms of sample size. Specifically, the CORD comprises 1.1K receipts and 30 semantic labels (e.g., menu name, price, total price, etc.), whereas the Form Understanding (FUNSD) dataset only includes 200 forms.

In line with prior research, we approach this task as a sequence labeling problem using BIO tags. Specifically, the GenDoc encoder is fed the instruction, text, layout, and image in the same manner as the Text Infilling task. The decoder input consists of the instruction, text, and position embedding sequences, which will be processed by the textual expert. With officially provided images and OCR annotations, the model is fine-tuned for 50 epochs with a batch size of 4. 

In alignment with previous studies, such as \cite{appalaraju2021docformer,powalski2021going,huang2022layoutlmv3}, the entity level F1 score for the test set is presented in Table \ref{tab:final_result} for comparative analysis. 
The results demonstrate that our proposed model exhibits a slight improvement in performance compared to the state-of-the-art method.

\paragraph{Document Classification}

Document classification involves predicting the appropriate category for a given document. To evaluate the performance of our model, we utilize the RVL-CDIP dataset. Our model's input is consistent with the pre-training process, which involves providing a task-specific instruction: ``\textit{What is the category of the document?}'' along with the text, image, and layout information obtained from an off-the-shelf OCR.
In line with the approach adopted in \cite{lewis2020bart} where a seq2seq model is utilized for classification, we feed the document text into the decoder and perform classification on the hidden state of the <eos> token. The model is fine-tuned for 10 epochs with a batch size of 96.

We report the accuracy of the classification on the test data in table \ref{tab:final_result}. The result shows our model has a desirable performance in capturing the coarse information from text and images of a document.

 \begin{table*}
	\centering
	\resizebox{0.75\linewidth}{!}{
		\begin{tabular} {llcccccc} 
         \toprule
			&  \bf Task & GenDoc & \textit{w/o} TI & \textit{w/o} CP& \textit{w/o} ITP${^\dagger}$ & \textit{w/o} MoE & \textit{w/o} Rel$^\ast$ \\
		 \midrule
			& DocVQA: ANLS & 78.15 & 64.87 & 77.98 & - & 77.87 & 75.17 \\
			& PubLaynet: mAP & 92.31 & 92.83 & 90.15 & - & 91.60 & 91.06 \\
          \bottomrule
		\end{tabular}}
	\caption{Ablation study of GenDoc: Using the pre-trained model, we fine-tuned the model on the DocVQA and the PubLayNet dataset for 10 epochs, respectively. $\dagger$ Deleting image token prediction task alone would cause Nan loss problem during pre-training. $\ast$ Rel means the disentangled relative attention scheme. }
    \label{tab:ablation}
\end{table*}

 \begin{table*}
	\centering
	\resizebox{0.68\linewidth}{!}{
		\begin{tabular} {llcccc} 
         \toprule
            & \multirow{2}{*}
            {\bf Model \& OCR} &  LayoutLMv3 & LayoutLMv3 & GenDoc & GenDoc\\
			& & Original OCR & MS OCR & Original OCR & MS OCR \\
		 \midrule
			& DocVQA: ANLS & 68.5 & 73.2 & 73.5  & 77.5 \\
          \bottomrule
			
		\end{tabular}
	}
	\caption{Experimental results on the effectiveness of OCR. All experiments are conducted using the train dataset of DocVQA and report results on the validation dataset. Original OCR refers to the officially provided OCR results from DocVQA, and MS OCR refers to the Microsoft Read API service.}
    \label{tab:ablation_ocr}
\end{table*}

\subsection{Ablation Study}
\label{sec:ablation}
In order to maintain a balance between computational efficiency and experimental validity, all ablation experiments are executed on a single node utilizing 8 A100 GPUs. To ensure comparability across all pre-training tasks, the ratio of batch size is kept constant, as previously discussed in the methodology. Additionally, each model is pre-trained for 100k steps under identical conditions to maintain a consistent experimental setup.

\paragraph{Pre-training}
This section aims to investigate the individual contribution of each pre-training task to downstream tasks. As previously discussed, three pre-training tasks are introduced for each modality (text, image, and layout) respectively and the results of unified training are reported. In order to gain a deeper understanding of the effect of each pre-training task, additional experiments are designed by removing each pre-training task individually and evaluating the model's performance on downstream tasks. The results of these experiments are presented in Table \ref{tab:ablation}, where two downstream tasks are selected as representative examples of text and image-layout modality tasks.

As observed from the results presented in Table \ref{tab:ablation}, the exclusion of text pre-training (Text Infilling, TI) results in a significant decline in performance for the DocVQA task (-$13.28$ ANLS). Conversely, a slight improvement in performance is observed for the PubLayNet task (+$0.52$ mAP), which is consistent with the findings reported in~\cite{wang2022ofa}. Additionally, the removal of layout pre-training task~(CP) results in only a moderate decrease in performance for the DocVQA task, indicating that DocVQA is a text-centric task.
The results for the object detection task, PubLaynet, reveal that the masked Coordinate Prediction task (CP) is crucial, which is consistent with the design consideration. This is due to the fact that the CP task not only trains the layout modality but also functions as an unsupervised object detection task, thus the removal of this task results in a substantial decrease in performance~(-$2.16$ mAP).
It should be noted that, upon the removal of the image token prediction task (ITP) alone, the pre-training becomes unstable and convergence becomes difficult to achieve.
This phenomenon was consistently observed across multiple replication of this experiment. Through analysis of the training process, we conclude that this instability is likely due to the absence of a link between text and layout modality, which further highlights the importance of the masked Image Token Prediction~(ITP) task.

\paragraph{Model}
In our study, we conducted experiments to investigate the effects of specialized designs implemented in the encoder and decoder. Specifically, we incorporate disentangled attention into the encoder to enhance the representation of 2D positional information, as the order of text in a document is not inherently sequential. Through the removal of the relative disentangled attention component, we observed a significant decline in performance for both tasks~(-$2.98$ ANLS, -$1.25$ mAP), thereby demonstrating the effectiveness of this design. Additionally, we examined the impact of incorporating MoE, which is intended to mitigate confusion caused by the different modalities in our design. Different from previous work, the results are unexpectedly not significant, with only a slight increase in performance observed upon the inclusion of MoE~(+$0.28$ ANLS, +$0.71$ mAP).
In our analysis, we think this phenomenon is a result of the level of training our downstream tasks received. Unlike the few-shot paradigm utilized in previous studies ~\cite{wang2022beitv3}, we use a fully-supervised fine-tuning approach with the full dataset, which likely contributes to the observed performance proximity.

\paragraph{OCR}
In this section, we evaluate the impact of OCR effects on performance. In various tasks related to document understanding, textual input plays a crucial role in determining performance. However, many previous studies have utilized internal OCR techniques and have not made the parsed data available. Therefore, we conduct experiments to assess the significance of OCR quality. Specifically, we test the pre-trained model using different OCR results on the DocVQA task, which heavily relies on textual inputs. The results of these experiments are presented in Table \ref{tab:ablation_ocr}. 

We replicated LayoutLMv3 using their open-sourced pre-trained checkpoint and code, and trained LayoutLMv3 model for 40 epochs. Additionally, we trained our GenDoc model using the pre-trained checkpoint for 10 epochs for each test. The results of our experimentation indicate that OCR quality has a substantial impact on performance. Furthermore, when utilizing the same quality of textual inputs, our model exhibits superior performance in comparison to previous state-of-the-art work. This further supports the assertion that our model is less susceptible to the negative effects of imperfect OCR.

\section{Related Work}
Our work is mostly related to recent work on multi-modal transformers for visual document understanding.
\citet{xu2020layoutlm} proposed the LayoutLM model to first combine the textual and layout information in a pre-trained transformer model. 
Inspired by BERT~\cite{devlin2019bert}, they designed the 2-D position embedding to denote the position of a token within a document.
They achieved much better performance compared with pure text-based approaches (e.g., BERT and Roberta~\cite{liu2019roberta}) on the downstream tasks (i.e., sequence labeling and document classification).
Following research efforts can be broadly categorized by the architecture and the pre-training objective design. 
\citet{li2021structurallm} proposed to use the cell-level positions rather than word-level positions (i.e., LayoutLM) to better understand the layout in documents. 
Various spatial-aware attention mechanisms~\cite{xu2021layoutlmv2,powalski2021going,appalaraju2021docformer,wang2022lilt} are proposed to capture layout information.

Pre-training strategy focuses on designing a pre-training objective to enable the model rely on cross-modality interaction.
Specifically, pre-training losses such as text-image alignment and text-image matching are essential to strengthen the tie between the visual and textual information~\cite{xu2021layoutlmv2,cho2021unifying,huang2022layoutlmv3}.
In addition, the choice of image backbone also makes a difference as it would affect the ability to understand the document image and  the pre-training strategy for the visual modality. 
While most literature adopts different variants of ResNet~\cite{he2016deep} to for down sampling and obtain a flattern sequence to the transformer,  LayoutLMv3~\cite{huang2022layoutlmv3} achieves state-of-the-art performance by using image patches.
Such a design also allows them to perform masked image token prediction.
However, the input sequence length could be too long to include all the information. 
Our approach mitigate the above limitation by combining the usage of ResNet and sequence-to-sequence architecture.

\section{Conclusions and Future Work}
The GenDoc model is an integrated sequence-to-sequence architecture for document understanding. To achieve this, We design text infilling, masked image token prediction, and masked coordinate prediction pre-training objectives for the textual, visual, and layout modality, respectively. The sequence-to-sequence model design allows for fine-tuning and inference on all downstream tasks utilizing the same structure.
A comprehensive examination of the proposed approach was conducted through a series of experiments on four standard document understanding tasks. The results of these experiments demonstrate the effectiveness of the proposed approach. Additionally, it was found that the proposed approach yields more robust performance in the presence of imperfect OCR when compared to encoder-only models.

Our future work includes extending our model to the scenario where OCR is not required, increasing both model size and data size to enable zero-shot prediction capabilities, and integrating the vision-language models across various domains such as visual document~\cite{mathew2021docvqa}, Infographics~\cite{mathew2022infographicvqa}, webpage image~\cite{tanaka2021visualmrc,chen2021websrc}, and visual scene understanding~\cite{lin2014microsoft}. 

\newpage

\bibliography{anthology,custom}

\begin{thebibliography}{47}
\expandafter\ifx\csname natexlab\endcsname\relax\def\natexlab#1{#1}\fi

\bibitem[{Appalaraju et~al.(2021)Appalaraju, Jasani, Kota, Xie, and
  Manmatha}]{appalaraju2021docformer}
Srikar Appalaraju, Bhavan Jasani, Bhargava~Urala Kota, Yusheng Xie, and
  R~Manmatha. 2021.
\newblock \href {https://arxiv.org/abs/2106.11539} {Docformer: End-to-end
  transformer for document understanding}.
\newblock In \emph{Proceedings of the IEEE/CVF International Conference on
  Computer Vision}.

\bibitem[{Bao et~al.(2021)Bao, Dong, Piao, and Wei}]{bao2021beit}
Hangbo Bao, Li~Dong, Songhao Piao, and Furu Wei. 2021.
\newblock \href {https://arxiv.org/abs/2106.08254} {Beit: Bert pre-training of
  image transformers}.
\newblock In \emph{Proceedings of International Conference on Learning
  Representations}.

\bibitem[{Cai and Vasconcelos(2018)}]{cai2018cascade}
Zhaowei Cai and Nuno Vasconcelos. 2018.
\newblock \href {https://arxiv.org/abs/1712.00726} {Cascade r-cnn: Delving into
  high quality object detection}.
\newblock In \emph{Proceedings of CVPR}.

\bibitem[{Chen et~al.(2022)Chen, Saxena, Li, Fleet, and
  Hinton}]{chen2022pix2seq}
Ting Chen, Saurabh Saxena, Lala Li, David~J. Fleet, and Geoffrey Hinton. 2022.
\newblock \href {https://arxiv.org/pdf/2109.10852.pdf} {Pix2seq: A language
  modeling framework for object detection}.
\newblock In \emph{Proceedings of ICLR}.

\bibitem[{Chen et~al.(2021)Chen, Zhao, Chen, Ji, Zhang, Luo, Xiong, and
  Yu}]{chen2021websrc}
Xingyu Chen, Zihan Zhao, Lu~Chen, JiaBao Ji, Danyang Zhang, Ao~Luo, Yuxuan
  Xiong, and Kai Yu. 2021.
\newblock \href {https://arxiv.org/pdf/2101.09465.pdf} {Websrc: A dataset for
  web-based structural reading comprehension}.
\newblock In \emph{Proceedings of the 2021 Conference on Empirical Methods in
  Natural Language Processing}.

\bibitem[{Chen et~al.(2020)Chen, Li, Yu, El~Kholy, Ahmed, Gan, Cheng, and
  Liu}]{chen2020uniter}
Yen-Chun Chen, Linjie Li, Licheng Yu, Ahmed El~Kholy, Faisal Ahmed, Zhe Gan,
  Yu~Cheng, and Jingjing Liu. 2020.
\newblock \href {https://arxiv.org/pdf/1909.11740.pdf} {Uniter: Universal
  image-text representation learning}.
\newblock In \emph{Proceedings of ECCV}.

\bibitem[{Cho et~al.(2021)Cho, Lei, Tan, and Bansal}]{cho2021unifying}
Jaemin Cho, Jie Lei, Hao Tan, and Mohit Bansal. 2021.
\newblock \href {http://proceedings.mlr.press/v139/cho21a/cho21a.pdf} {Unifying
  vision-and-language tasks via text generation}.
\newblock In \emph{Proceedings of International Conference on Machine
  Learning}.

\bibitem[{Devlin et~al.(2019)Devlin, Chang, Lee, and
  Toutanova}]{devlin2019bert}
Jacob Devlin, Ming-Wei Chang, Kenton Lee, and Kristina Toutanova. 2019.
\newblock \href {https://arxiv.org/abs/1810.04805} {Bert: Pre-training of deep
  bidirectional transformers for language understanding}.
\newblock In \emph{Proceedings of NAACL}.

\bibitem[{Gao et~al.(2019)Gao, Huang, D{\'e}jean, Meunier, Yan, Fang, Kleber,
  and Lang}]{gao2019icdar}
Liangcai Gao, Yilun Huang, Herv{\'e} D{\'e}jean, Jean-Luc Meunier, Qinqin Yan,
  Yu~Fang, Florian Kleber, and Eva Lang. 2019.
\newblock \href {https://ieeexplore.ieee.org/document/8978120} {Icdar 2019
  competition on table detection and recognition (ctdar)}.
\newblock In \emph{Proceedings of International Conference on Document Analysis
  and Recognition}.

\bibitem[{Guillaume~Jaume(2019)}]{jaume2019funsd}
Jean-Philippe~Thiran Guillaume~Jaume, Hazim Kemal~Ekenel. 2019.
\newblock \href {https://arxiv.org/abs/1905.13538} {{FUNSD}: A dataset for form
  understanding in noisy scanned documents}.
\newblock In \emph{Proceedings of International Conference on Document Analysis
  and Recognition Workshops}.

\bibitem[{Harley et~al.(2015{\natexlab{a}})Harley, Ufkes, and
  Derpanis}]{harley2015evaluation}
Adam~W Harley, Alex Ufkes, and Konstantinos~G Derpanis. 2015{\natexlab{a}}.
\newblock \href {https://arxiv.org/abs/1502.07058} {Evaluation of deep
  convolutional nets for document image classification and retrieval}.
\newblock In \emph{Proceedings of International Conference on Document Analysis
  and Recognition}.

\bibitem[{Harley et~al.(2015{\natexlab{b}})Harley, Ufkes, and
  Derpanis}]{adam2015cdip}
Adam~W. Harley, Alex Ufkes, and Konstantinos~G. Derpanis. 2015{\natexlab{b}}.
\newblock \href {https://arxiv.org/pdf/1502.07058v1.pdf} {Evaluation of deep
  convolutional nets for document image classification and retrieval}.
\newblock In \emph{2015 13th International Conference on Document Analysis and
  Recognition (ICDAR)}.

\bibitem[{He et~al.(2022)He, Chen, Xie, Li, Doll{\'a}r, and
  Girshick}]{he2022masked}
Kaiming He, Xinlei Chen, Saining Xie, Yanghao Li, Piotr Doll{\'a}r, and Ross
  Girshick. 2022.
\newblock \href
  {https://openaccess.thecvf.com/content/CVPR2022/html/He_Masked_Autoencoders_Are_Scalable_Vision_Learners_CVPR_2022_paper.html}
  {Masked autoencoders are scalable vision learners}.
\newblock In \emph{Proceedings of the IEEE/CVF Conference on Computer Vision
  and Pattern Recognition}.

\bibitem[{He et~al.(2016)He, Zhang, Ren, and Sun}]{he2016deep}
Kaiming He, Xiangyu Zhang, Shaoqing Ren, and Jian Sun. 2016.
\newblock \href
  {https://openaccess.thecvf.com/content_cvpr_2016/papers/He_Deep_Residual_Learning_CVPR_2016_paper.pdf}
  {Deep residual learning for image recognition}.
\newblock In \emph{Proceedings of the IEEE conference on computer vision and
  pattern recognition}.

\bibitem[{He et~al.(2021)He, Liu, Gao, and Chen}]{he2021deberta}
Pengcheng He, Xiaodong Liu, Jianfeng Gao, and Weizhu Chen. 2021.
\newblock \href {https://arxiv.org/abs/2006.03654} {Deberta: Decoding-enhanced
  bert with disentangled attention}.
\newblock In \emph{Proceedings of ICLR}.

\bibitem[{Huang et~al.(2022)Huang, Lv, Cui, Lu, and Wei}]{huang2022layoutlmv3}
Yupan Huang, Tengchao Lv, Lei Cui, Yutong Lu, and Furu Wei. 2022.
\newblock \href {https://arxiv.org/abs/2204.08387} {Layoutlmv3: Pre-training
  for document ai with unified text and image masking}.
\newblock \emph{arXiv preprint arXiv:2204.08387}.

\bibitem[{Hwang et~al.(2021)Hwang, Lee, Yim, Kim, and Seo}]{hwang2021cost}
Wonseok Hwang, Hyunji Lee, Jinyeong Yim, Geewook Kim, and Minjoon Seo. 2021.
\newblock \href {https://arxiv.org/abs/2104.08041} {Cost-effective end-to-end
  information extraction for semi-structured document images}.
\newblock In \emph{Proceedings of the 2021 Conference on Empirical Methods in
  Natural Language Processing}.

\bibitem[{Lewis et~al.(2006)Lewis, Agam, Argamon, Frieder, Grossman, and
  Heard}]{david2006iit}
D.~Lewis, G.~Agam, S.~Argamon, O.~Frieder, D.~Grossman, and J.~Heard. 2006.
\newblock \href {https://dl.acm.org/doi/10.1145/1148170.1148307} {Building a
  test collection for complex document information processing}.
\newblock In \emph{Proceedings of the 29th annual international ACM SIGIR
  conference on Research and development in information retrieval}.

\bibitem[{Lewis et~al.(2020)Lewis, Liu, Goyal, Ghazvininejad, Mohamed, Levy,
  Stoyanov, and Zettlemoyer}]{lewis2020bart}
Mike Lewis, Yinhan Liu, Naman Goyal, Marjan Ghazvininejad, Abdelrahman Mohamed,
  Omer Levy, Veselin Stoyanov, and Luke Zettlemoyer. 2020.
\newblock \href {https://aclanthology.org/2020.acl-main.} {Bart: Denoising
  sequence-to-sequence pre-training for natural language generation,
  translation, and comprehension}.
\newblock In \emph{Proceedings of the 58th Annual Meeting of the Association
  for Computational Linguistics}.

\bibitem[{Li et~al.(2021)Li, Bi, Yan, Wang, Huang, Huang, and
  Si}]{li2021structurallm}
Chenliang Li, Bin Bi, Ming Yan, Wei Wang, Songfang Huang, Fei Huang, and Luo
  Si. 2021.
\newblock \href {https://arxiv.org/pdf/2105.11210.pdf} {Structurallm:
  Structural pre-training for form understanding}.
\newblock In \emph{Proceedings of ACL}.

\bibitem[{Li et~al.(2022)Li, Xu, Lv, Cui, Zhang, and Wei}]{Li2022dit}
Junlong Li, Yiheng Xu, Tengchao Lv, Lei Cui, Cha Zhang, and Furu Wei. 2022.
\newblock \href {https://arxiv.org/pdf/2203.02378.pdf} {Dit: Self-supervised
  pre-training for document image transformer}.
\newblock In \emph{Proceedings of the 30th ACM International Conference on
  Multimedia}.

\bibitem[{Lin et~al.(2017)Lin, Doll{\'a}r, Girshick, He, Hariharan, and
  Belongie}]{lin2017fpn}
Tsung-Yi Lin, Piotr Doll{\'a}r, Ross Girshick, Kaiming He, Bharath Hariharan,
  and Serge Belongie. 2017.
\newblock \href {https://arxiv.org/abs/1612.03144} {Feature pyramid networks
  for object detection}.
\newblock In \emph{Proceedings of CVPR}.

\bibitem[{Lin et~al.(2014)Lin, Maire, Belongie, Hays, Perona, Ramanan,
  Doll{\'a}r, and Zitnick}]{lin2014microsoft}
Tsung-Yi Lin, Michael Maire, Serge Belongie, James Hays, Pietro Perona, Deva
  Ramanan, Piotr Doll{\'a}r, and C~Lawrence Zitnick. 2014.
\newblock \href {https://arxiv.org/abs/1405.0312} {Microsoft coco: Common
  objects in context}.
\newblock In \emph{Proceedings of ECCV}.

\bibitem[{Liu et~al.(2019)Liu, Ott, Goyal, Du, Joshi, Chen, Levy, Lewis,
  Zettlemoyer, and Stoyanov}]{liu2019roberta}
Yinhan Liu, Myle Ott, Naman Goyal, Jingfei Du, Mandar Joshi, Danqi Chen, Omer
  Levy, Mike Lewis, Luke Zettlemoyer, and Veselin Stoyanov. 2019.
\newblock \href {https://arxiv.org/abs/1907.11692} {Roberta: A robustly
  optimized bert pretraining approach}.
\newblock \emph{arXiv preprint arXiv:1907.11692}.

\bibitem[{Mathew et~al.(2022)Mathew, Bagal, Tito, Karatzas, Valveny, and
  Jawahar}]{mathew2022infographicvqa}
Minesh Mathew, Viraj Bagal, Rub{\`e}n Tito, Dimosthenis Karatzas, Ernest
  Valveny, and CV~Jawahar. 2022.
\newblock \href
  {https://openaccess.thecvf.com/content/WACV2022/papers/Mathew_InfographicVQA_WACV_2022_paper.pdf}
  {Infographicvqa}.
\newblock In \emph{Proceedings of the IEEE/CVF Winter Conference on
  Applications of Computer Vision}.

\bibitem[{Mathew et~al.(2021)Mathew, Karatzas, and Jawahar}]{mathew2021docvqa}
Minesh Mathew, Dimosthenis Karatzas, and CV~Jawahar. 2021.
\newblock \href {https://arxiv.org/abs/2007.00398} {Docvqa: A dataset for vqa
  on document images}.
\newblock In \emph{Proceedings of the IEEE/CVF winter conference on
  applications of computer vision}.

\bibitem[{Neumann and Matas(2012)}]{neumann2012real}
Luk{\'a}{\v{s}} Neumann and Ji{\v{r}}{\'\i} Matas. 2012.
\newblock \href {https://ieeexplore.ieee.org/document/6248097} {Real-time scene
  text localization and recognition}.
\newblock In \emph{Proceedings of CVPR}.

\bibitem[{Park et~al.(2019)Park, Shin, Lee, Lee, Surh, Seo, and
  Lee}]{Park2019cord}
Seunghyun Park, Seung Shin, Bado Lee, junyeop Lee, Jaeheung Surh, Minjoon Seo,
  and Hwalsuk Lee. 2019.
\newblock \href {https://openreview.net/pdf?id=SJl3z659UH} {Cord: A
  consolidated receipt dataset for post-ocr parsing}.
\newblock In \emph{Proceedings of NeurIPS}.

\bibitem[{Peng et~al.(2022)Peng, Pan, Wang, Luo, Zhang, Huang, Hu, Yin, Chen,
  Zhang et~al.}]{peng2022ernie}
Qiming Peng, Yinxu Pan, Wenjin Wang, Bin Luo, Zhenyu Zhang, Zhengjie Huang,
  Teng Hu, Weichong Yin, Yongfeng Chen, Yin Zhang, et~al. 2022.
\newblock \href {https://arxiv.org/abs/2210.06155} {Ernie-layout: Layout
  knowledge enhanced pre-training for visually-rich document understanding}.
\newblock In \emph{Findings of EMNLP}.

\bibitem[{Powalski et~al.(2021)Powalski, Borchmann, Jurkiewicz, Dwojak,
  Pietruszka, and Pa{\l}ka}]{powalski2021going}
Rafa{\l} Powalski, {\L}ukasz Borchmann, Dawid Jurkiewicz, Tomasz Dwojak,
  Micha{\l} Pietruszka, and Gabriela Pa{\l}ka. 2021.
\newblock \href
  {https://link.springer.com/chapter/10.1007/978-3-030-86331-9_47} {Going
  full-tilt boogie on document understanding with text-image-layout
  transformer}.
\newblock In \emph{Proceedings of International Conference on Document Analysis
  and Recognition}.

\bibitem[{Raffel et~al.(2020)Raffel, Shazeer, Roberts, Lee, Narang, Matena,
  Zhou, Li, and Liu}]{raffel2020exploring}
Colin Raffel, Noam Shazeer, Adam Roberts, Katherine Lee, Sharan Narang, Michael
  Matena, Yanqi Zhou, Wei Li, and Peter~J Liu. 2020.
\newblock \href {https://www.jmlr.org/papers/volume21/20-074/20-074.pdf}
  {Exploring the limits of transfer learning with a unified text-to-text
  transformer}.
\newblock \emph{Journal of Machine Learning Research}, 21:1--67.

\bibitem[{Ramesh et~al.(2022)Ramesh, Dhariwal, Nichol, Chu, and
  Chen}]{ramesh2022hierarchical}
Aditya Ramesh, Prafulla Dhariwal, Alex Nichol, Casey Chu, and Mark Chen. 2022.
\newblock \href {https://arxiv.org/pdf/2204.06125.pdf} {Hierarchical
  text-conditional image generation with clip latents}.
\newblock \emph{arXiv preprint arXiv:2204.06125}.

\bibitem[{Ren et~al.(2015)Ren, He, Girshick, and Sun}]{ren2015fasterrcnn}
Shaoqing Ren, Kaiming He, Ross Girshick, and Jian Sun. 2015.
\newblock \href
  {https://proceedings.neurips.cc/paper/2015/hash/14bfa6bb14875e45bba028a21ed38046-Abstract.html}
  {Faster r-cnn: Towards real-time object detection with region proposal
  networks}.
\newblock In \emph{Proceedings of NeurIPS}.

\bibitem[{Su et~al.(2020)Su, Zhu, Cao, Li, Lu, Wei, and Dai}]{su2020vl}
Weijie Su, Xizhou Zhu, Yue Cao, Bin Li, Lewei Lu, Furu Wei, and Jifeng Dai.
  2020.
\newblock \href {https://arxiv.org/pdf/1908.08530.pdf} {Vl-bert: Pre-training
  of generic visual-linguistic representations}.
\newblock In \emph{Proceedings of International Conference on Learning
  Representations}.

\bibitem[{Tan and Bansal(2019)}]{tan2019lxmert}
Hao Tan and Mohit Bansal. 2019.
\newblock \href {https://arxiv.org/pdf/1908.07490.pdf} {Lxmert: Learning
  cross-modality encoder representations from transformers}.
\newblock In \emph{Proceedings of the 2019 Conference on Empirical Methods in
  Natural Language Processing and the 9th International Joint Conference on
  Natural Language Processing (EMNLP-IJCNLP)}.

\bibitem[{Tanaka et~al.(2021)Tanaka, Nishida, and
  Yoshida}]{tanaka2021visualmrc}
Ryota Tanaka, Kyosuke Nishida, and Sen Yoshida. 2021.
\newblock \href
  {https://ojs.aaai.org/index.php/AAAI/article/download/17635/17442}
  {Visualmrc: Machine reading comprehension on document images}.
\newblock In \emph{Proceedings of the AAAI Conference on Artificial
  Intelligence}.

\bibitem[{Van Den~Oord et~al.(2017)Van Den~Oord, Vinyals
  et~al.}]{van2017neural}
Aaron Van Den~Oord, Oriol Vinyals, et~al. 2017.
\newblock \href
  {https://papers.nips.cc/paper/2017/file/7a98af17e63a0ac09ce2e96d03992fbc-Paper.pdf}
  {Neural discrete representation learning}.
\newblock In \emph{Proceedings of NeurIPS}.

\bibitem[{Vaswani et~al.(2017)Vaswani, Shazeer, Parmar, Uszkoreit, Jones,
  Gomez, Kaiser, and Polosukhin}]{vaswani2017attention}
Ashish Vaswani, Noam Shazeer, Niki Parmar, Jakob Uszkoreit, Llion Jones,
  Aidan~N Gomez, {\L}ukasz Kaiser, and Illia Polosukhin. 2017.
\newblock \href {https://arxiv.org/abs/1706.03762} {Attention is all you need}.
\newblock In \emph{Proceedings of NeurIPS}.

\bibitem[{Wang et~al.(2022{\natexlab{a}})Wang, Jin, and Ding}]{wang2022lilt}
Jiapeng Wang, Lianwen Jin, and Kai Ding. 2022{\natexlab{a}}.
\newblock \href {https://aclanthology.org/2022.acl-long.534/} {Lilt: A simple
  yet effective language-independent layout transformer for structured document
  understanding}.
\newblock In \emph{Proceedings of the 60th Annual Meeting of the Association
  for Computational Linguistics (Volume 1: Long Papers)}.

\bibitem[{Wang et~al.(2022{\natexlab{b}})Wang, Yang, Men, Lin, Bai, Li, Ma,
  Zhou, Zhou, and Yang}]{wang2022ofa}
Peng Wang, An~Yang, Rui Men, Junyang Lin, Shuai Bai, Zhikang Li, Jianxin Ma,
  Chang Zhou, Jingren Zhou, and Hongxia Yang. 2022{\natexlab{b}}.
\newblock \href {https://arxiv.org/pdf/2202.03052.pdf} {Ofa: Unifying
  architectures, tasks, and modalities through a simple sequence-to-sequence
  learning framework}.
\newblock In \emph{Proceedings of International Conference on Machine
  Learning}. PMLR.

\bibitem[{Wang et~al.(2022{\natexlab{c}})Wang, Bao, Dong, Bjorck, Peng, Liu,
  Aggarwal, Mohammed, Singhal, Som, and Wei}]{wang2022beitv3}
Wenhui Wang, Hangbo Bao, Li~Dong, Johan Bjorck, Zhiliang Peng, Qiang Liu, Kriti
  Aggarwal, Owais~Khan Mohammed, Saksham Singhal, Subhojit Som, and Furu Wei.
  2022{\natexlab{c}}.
\newblock \href {http://arxiv.org/abs/arXiv:2208.10442} {Image as a foreign
  language: Beit pretraining for all vision and vision-language tasks}.

\bibitem[{Wang et~al.(2021)Wang, Bao, Dong, and Wei}]{wang2021vlmo}
Wenhui Wang, Hangbo Bao, Li~Dong, and Furu Wei. 2021.
\newblock \href {https://arxiv.org/abs/2111.02358} {Vlmo: Unified
  vision-language pre-training with mixture-of-modality-experts}.
\newblock \emph{arXiv preprint arXiv:2111.02358}.

\bibitem[{Xie et~al.(2022)Xie, Zhang, Cao, Lin, Bao, Yao, Dai, and
  Hu}]{xie2022simmim}
Zhenda Xie, Zheng Zhang, Yue Cao, Yutong Lin, Jianmin Bao, Zhuliang Yao,
  Qi~Dai, and Han Hu. 2022.
\newblock \href
  {https://openaccess.thecvf.com/content/CVPR2022/html/Xie_SimMIM_A_Simple_Framework_for_Masked_Image_Modeling_CVPR_2022_paper.html}
  {Simmim: A simple framework for masked image modeling}.
\newblock In \emph{Proceedings of CVPR}.

\bibitem[{Xu et~al.(2021)Xu, Xu, Lv, Cui, Wei, Wang, Lu, Florencio, Zhang, Che
  et~al.}]{xu2021layoutlmv2}
Yang Xu, Yiheng Xu, Tengchao Lv, Lei Cui, Furu Wei, Guoxin Wang, Yijuan Lu,
  Dinei Florencio, Cha Zhang, Wanxiang Che, et~al. 2021.
\newblock \href {https://arxiv.org/abs/2012.14740} {Layoutlmv2: Multi-modal
  pre-training for visually-rich document understanding}.
\newblock In \emph{Proceedings of the 59th Annual Meeting of the Association
  for Computational Linguistics and the 11th International Joint Conference on
  Natural Language Processing (Volume 1: Long Papers)}.

\bibitem[{Xu et~al.(2020)Xu, Li, Cui, Huang, Wei, and Zhou}]{xu2020layoutlm}
Yiheng Xu, Minghao Li, Lei Cui, Shaohan Huang, Furu Wei, and Ming Zhou. 2020.
\newblock \href {https://arxiv.org/abs/1912.13318} {Layoutlm: Pre-training of
  text and layout for document image understanding}.
\newblock In \emph{Proceedings of the 26th ACM SIGKDD International Conference
  on Knowledge Discovery \& Data Mining}.

\bibitem[{Yu et~al.(2022)Yu, Xu, Koh, Luong, Baid, Wang, Vasudevan, Ku, Yang,
  Ayan et~al.}]{yu2022scaling}
Jiahui Yu, Yuanzhong Xu, Jing~Yu Koh, Thang Luong, Gunjan Baid, Zirui Wang,
  Vijay Vasudevan, Alexander Ku, Yinfei Yang, Burcu~Karagol Ayan, et~al. 2022.
\newblock \href {https://arxiv.org/abs/2206.10789} {Scaling autoregressive
  models for content-rich text-to-image generation}.
\newblock \emph{arXiv preprint arXiv:2206.10789}.

\bibitem[{Zhong et~al.(2019)Zhong, Tang, and Yepes}]{zhong2019publaynet}
Xu~Zhong, Jianbin Tang, and Antonio~Jimeno Yepes. 2019.
\newblock \href {https://arxiv.org/abs/1908.07836} {Publaynet: largest dataset
  ever for document layout analysis}.
\newblock In \emph{Proceedings of International Conference on Document Analysis
  and Recognition}.

\end{thebibliography}
\bibliographystyle{acl_natbib}

\appendix
\section{Appendix}
\label{sec:appendix}
\subsection{Downstream tasks datasets characteristic}
The table \ref{tab: characteristic} reports the characteristic of the dataset for each downstream tasks.

\subsection{Hyper-parameters for downstream task}
In Table \ref{tab: hyperparameter}, a comprehensive report of all the hyperparameters utilized during fine-tuning of the pre-trained model for four distinct downstream tasks is presented for the purpose of reproducibility.

\subsection{More detail for experiments}
\paragraph{Pre-training}
The GenDoc base is a standard transformers consisting of a 6 layers encoder and 6 layers decoder with a hidden size of 768. While the Gendoc large model comprises a 12-layer encoder and 12-layer decoder with a hidden size of 1024. The learning rate for the main network during pre-training is $2e-5$, and $5e-5$ for the image backbone. The learning rate is linearly warmed up over 10,000 steps for the base model, and 30,000 steps for the large model.

\paragraph{DocVQA}
In the DocVQA dataset, some questions have multiple correct answers. We pairs each answer the corresponding document and question to form a sample, resulting in a final training data size of 56,259.  We truncate the tokenized document text when its length exceeds $800$, and concatenate the instruction, question, and document text to form the textual input. The document image is resized into $448\times448$ as the visual input. A label smoothing rate of $0.1$ was also used to prevent over-fitting. During inference, we use beam search with a beam size of $4$ and set the maximum generation length as $200$. 

\paragraph{Publaynet}
During the experiment, we apply the identical image augmentation strategy used in Layoutlmv3 and DiT, which randomly crop the image into a fixed size and then randomly resize it to a variant of scale. We also apply the sequence augmentation technique proposed in Pix2Seq~\cite{chen2022pix2seq} which is found to be helpful in our setting. During training, we generate synthetic noise through random scale and shifting the ground truth bounding box to construct a noise sequence [$\widetilde{x}_1, \widetilde{y}_1, \widetilde{x}_2, \widetilde{y}_2, class_{noise}$], where the $[\widetilde{x}_1, \widetilde{y}_1, \widetilde{x}_2, \widetilde{y}_2]$ is the synthetic bounding box, and $class_{noise}$ is the class to indicate the bounding box is noise. The noised sequence is appended to the end of ground truth objects to fill a maximum sequence length of 60 objects. The training objective is to predict the tokens for the ground truth sequences and distinguish whether the bounding box is a noise object.
We also increase the number of layout tokens from $1000$ to $2000$ for a higher dequantization precision. Specifically, the bounding box is quantized by $x_{quant} = round(x_{norm}\times1999)$, where $x_{norm}$ is the normalized bounding box and $x_{quant}$ is the quantized token. 

\paragraph{CORD}
To better align the attention between the decoder tokens and encoder output hidden states, we also add to each decoder input embedding their corresponding hidden state from the encoder output. Then, the decoder output hidden states corresponding to the textual tokens are passed through a classifier to predict its label. 

\paragraph{RVL-CDIP}
Our internal OCR parser is not able to process certain images, such as posters and handwriting letters, within the dataset. For these images, the OCR text input is left as an empty string. During the experiments, the tokenized document text is truncated when its length exceeds $512$. The images are resized to $448 \times 448$. Additionally, a label smoothing rate of $0.1$ is employed during the training process.

\begin{table*}
    \centering
    \resizebox{0.6\linewidth}{!}{
    \begin{tabular}{ccccc}
    \toprule
         &  \bf{Dataset} & \bf{Train} & \bf{Validation} & \bf{Test} \\
        \midrule
         &  DocVQA & 39,463 & 5,349 & 5,188 \\
         &  Publaynet & 335,703 & 11,245 & - \\
         &  CORD & 800 & 100 & 100 \\
         &  RVL-CDIP & 320,000 & 40,000 & 40,000 \\
    \bottomrule
    \end{tabular}
    }
    \caption{Datasets characteristic of four downstream tasks}
    \label{tab: characteristic}
\end{table*}
\begin{table*}
    \centering
    \resizebox{0.8\linewidth}{!}{
    \begin{tabular} {llcccccc} 
    \toprule
         & \bf{Dataset} &  \bf{Batch size} & \bf{Learning rate}  & \bf{Scheduler} & \bf{Epoch} & \bf{Warm-up} \\
         \midrule
           & DocVQA & 48 & 5e-5 & Cosine  & 10 & 1 epoch  \\
          \textsc{GenDoc} & Publaynet & 64 & 2e-4 & Linear & 50 & 1k steps \\
         \textsc{Base}  & CORD & 4 & 2e-5 & Linear & 50 & 1 epoch \\
           & RVL-CDIP & 96 & 2e-5 & Linear & 10 & 1 epoch \\
           \midrule\midrule
            & DocVQA & 48 & 2e-5 & Cosine  & 10 & 1 epoch  \\
         \textsc{GenDoc}  & Publaynet & - & - & - & - & - \\
          \textsc{Large} & CORD & 4 & 1e-5 & Linear & 50 & 1 epoch \\
           & RVL-CDIP & 96 & 1e-5 & Linear & 10 & 1 epoch \\
          \bottomrule
        \end{tabular}
    }
    \caption{Hyper-parameters used for four downstream tasks}
    \label{tab: hyperparameter}
\end{table*}

\end{document}